\documentclass[letterpaper, 10 pt, conference]{ieeeconf}  

\IEEEoverridecommandlockouts                              

\usepackage{graphics} 
\usepackage{graphicx}
\usepackage{cite}
\usepackage{bbm}
\usepackage[breaklinks=true,bookmarks=true,colorlinks]{hyperref}
\usepackage{caption}
\usepackage{amsmath} 
\usepackage{amssymb}  
\usepackage[font=small]{caption}
\usepackage{multirow}
\usepackage{tabularx}
\usepackage{color, colortbl}

\definecolor{LightCyan}{rgb}{0.88,1,1}
\definecolor{lightgray}{rgb}{0.9,0.9,0.9}
\usepackage{color}

\title{\LARGE \bf
Elastic Motion Policy: An Adaptive Dynamical System \\ for Robust and Efficient One-Shot Imitation Learning
}

\author{Tianyu Li$^{*}$, Sunan Sun, Shubhodeep Shiv Aditya and 
Nadia Figueroa
\thanks{*Corresponding author. (e-mail: tianyuli@seas.upenn.edu)}
\thanks{All authors are with School of Engineering and Applied Science,
University of Pennsylvania, Pennsylvania, PA 19104 USA.}%
}

\begin{document}

\maketitle
\thispagestyle{empty}
\pagestyle{empty}

\begin{abstract}
Behavior cloning (BC) has become a staple imitation learning paradigm in robotics due to its ease of teaching robots complex skills directly from expert demonstrations. However, BC suffers from an inherent generalization issue. To solve this, the status quo solution is to gather more data. Yet, regardless of how much training data is available, out-of-distribution performance is still sub-par, lacks any formal guarantee of convergence and success, and is incapable of allowing and recovering from physical interactions with humans. These are critical flaws when robots are deployed in ever-changing human-centric environments. Thus, we propose Elastic Motion Policy (EMP), a one-shot imitation learning framework that allows robots to adjust their behavior based on the scene change while respecting the task specification. Trained from a single demonstration, EMP follows the dynamical systems paradigm where motion planning and control are governed by first-order differential equations with convergence guarantees. We leverage Laplacian editing in full end-effector space, $\mathbb{R}^3\times SO(3)$, and online convex learning of Lyapunov functions, to adapt EMP online to new contexts, avoiding the need to collect new demonstrations. We extensively validate our framework in real robot experiments, demonstrating its robust and efficient performance in dynamic environments, with obstacle avoidance and multi-step task capabilities. \url{https://elastic-motion-policy.github.io/EMP/}
\end{abstract}

\section{INTRODUCTION}
\vspace{-1.5pt}
As robots become more common in human-centric environments like homes, warehouses, hospitals, and factories, it is important to consider generating motion that allows for possible physical interactions with humans or other agents. Thus, a robot's behavior should be compliant and reactive. Yet, specifying robot motion with compliance and reactivity requires control and robotics expertise. Imitation Learning (IL) provides an intuitive way for specifying motion by learning from human demonstrations \cite{ravichandar2020recent}. However, even with demonstrations as guidance, such policies are difficult to define as unexpected physical interactions could lead the robot to Out-of-Distribution (OOD), which creates uncertainties in the robot behavior, leading to concern for both physical safety and perceived safety. One could collect more data to mitigate the OOD problem. However, accessing large data for dynamic human-centric environments may be intractable. More importantly, besides unexpected physical perturbations, the scenario could change, making a learned motion obsolete. The motion policy should be able to adapt to changes in the environment (even in real-time) and be reactive to newly presented physical perturbations and obstacles. Hence, the central research question that we tackle in this paper is: 

\textit{How do we allow robots to learn stable, compliant, adaptive and reactive behaviors from very few demonstrations?} 

We offer a solution to this problem by proposing the \textbf{Elastic Motion Policy (EMP)} framework.
\begin{figure}[!tbp]
  \centering
    \includegraphics[width=0.90\linewidth]{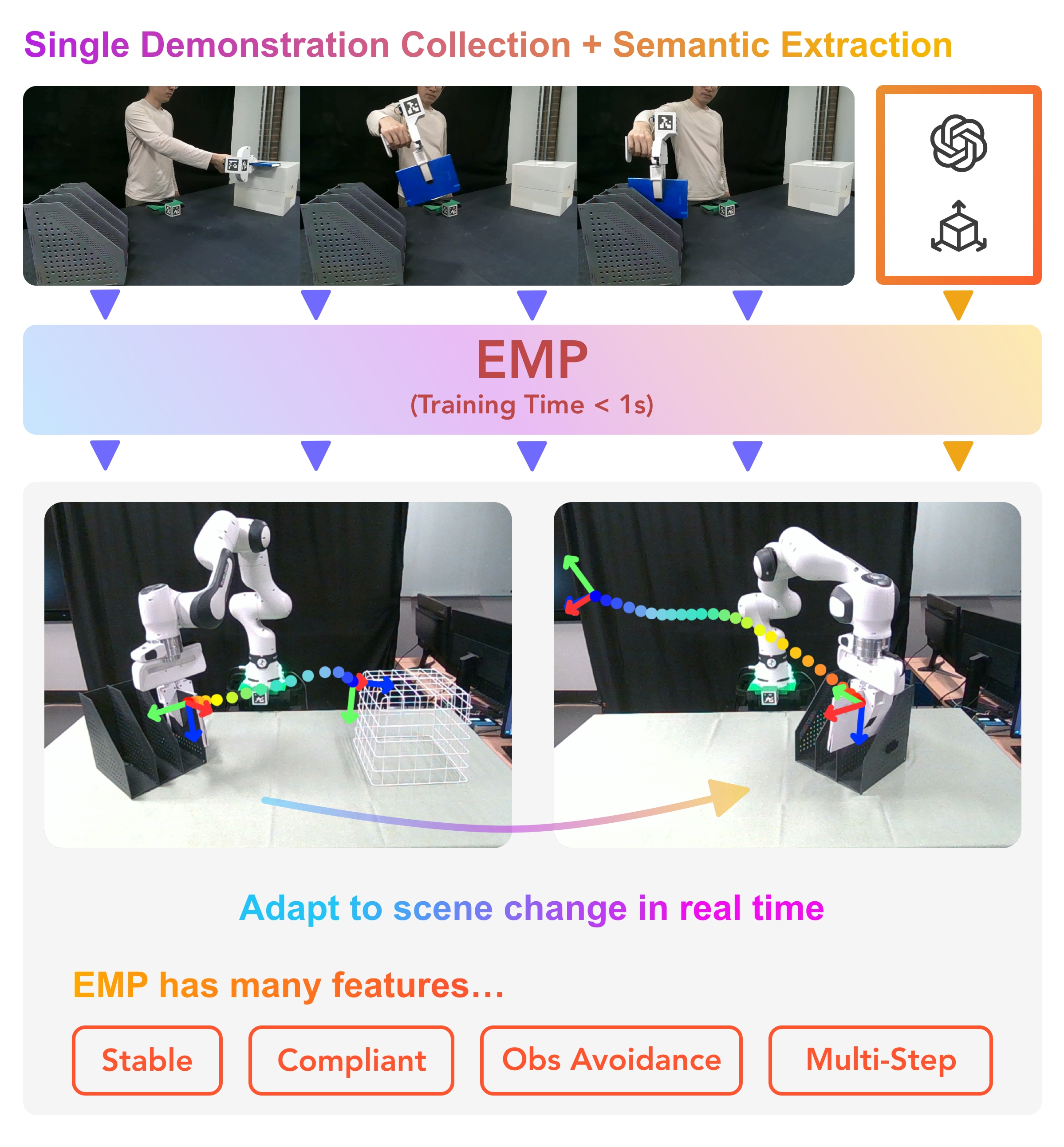}
  \caption{\textbf{EMP Framework Overview.} Given \textit{a single demonstration} and semantic knowledge of the scene, EMP can adjust a learned motion policy based on the changes in environment in real time. Additionally, EMP enables stable and compliant robotic control in scenarios involving obstacle avoidance and multi-step tasks.}
  \label{fig:overview}
  \vspace{-20pt}
\end{figure}
EMP builds upon the Dynamical System (DS) motion policy paradigm\cite{TEXTBOOK}: a behavior cloning (BC) paradigm that learns guaranteed stable motion policies and produces compliant and reactive robot behavior. Such properties are already addressing part of our desiderata. However, although DS motion policies can create a global motion that convergences towards an attractor or trajectory, such policies are not able to adapt to environmental changes without providing new demonstrations. This is, in fact, a caveat of any motion-level BC policy that only mimics the provided demonstrations.
We argue that a motion policy should not mimic demonstrations directly, but rather, extract relevant task information and adjust based on changes in the scene. In our prior work, Elastic-DS \cite{li2023task}, we addressed such a problem in the context of Gaussian Mixture Model (GMM) based Linear Parameter Varying DS (LPV-DS) motion policy learning \cite{figueroa2018physically,TEXTBOOK} by 1) extracting the GMM decomposition, which represents the complexity of a task, and 2) morphing the GMM to adjust to new geometric constraints based on the relevant objects in the environment during inference. While showing successful adaptation of DS motion policies on new contexts, Elastic-DS was formulated only in Euclidean space and required extra computation time for adaptation, taking minutes to adapt to a new context.

In this paper, we present a stable-guaranteed full-pose motion policy learned from very few (1-2) demonstrations that can adapt to new object configurations without new data at 30Hz, allowing for online adaptation even when the relevant task objects are moving. EMP is not just a standalone method that imitates demonstrations but rather a flexible framework that can combine many capabilities, such as task-level constraints, obstacle avoidance, and learning multi-step demonstrations. 
We assume 3D meshes of the objects of interest are available and can be used within an object tracking perception framework.
Our contributions are: 
\begin{enumerate}
    \item We extend the Elastic-DS\cite{li2023task} motion policy adaptation approach to orientation space by performing Laplacian editing on the quaternion manifold tangent plane.
    \item We propose a convex optimization formulation for full pose SE(3) LPV-DS learning \cite{se3lpvds}, which allows EMP to be updated in real time, approximately at 30Hz. 
    \item We design a pipeline to extract object keyposes from visual information as the EMP task parameters.
    \item We showcase the flexibility of the EMP framework to achieve real-time adaptation of motion policies, obstacle avoidance and learning of multi-step tasks.
\end{enumerate}

\noindent \textbf{Paper Organization} In Section \ref{sec:related} we summarize related works on IL frameworks. Section \ref{sec:background} introduces the backbone DS motion policy used to build EMP. Section \ref{sec:emp} introduces the EMP optimization steps while Section \ref{sec:pipeline} describes the training and inference pipeline. Section \ref{sec:experiment} showcases the real-world evaluations with varying complexity.
\section{RELATED WORK}
\label{sec:related}
Behavior Cloning (BC) \cite{osa2018algorithmic} has gained significant attention and has become a common approach for imitation learning (IL) \cite{ravichandar2020recent,chi2023diffusion, zhao2024aloha, florence2022implicit}. However, BC still suffers from compounding error and a lack of data to generalize for real-world deployment; it is necessary to consider safe and explainable behavior in human-centric applications, making a learned policy's robustness and stability crucial. Many efforts have been made towards robust and stable behavior cloning. 

A common approach for making an IL system more robust is by augmenting the training dataset. Methods following the DAgger paradigm \cite{pmlr-v15-ross11a-dagger, kelly2019hg, menda2019ensembledagger} require an online interactive expert, adding burden during training. To remedy workload, 
data augmentation techniques that inject noise during data collection have been proposed \cite{ke2021grasping, ke2023ccil,DART}.  
These methods, however, are not tractable when the environments are changing online and data inefficient.

An alternative approach is to pose constraints on the policy, which our proposed method follows. By applying methods drawing from control theory, a learned policy could be guided toward safe and stable regions from out-of-distribution cases. Stable-BC \cite{mehta2024stableBC} proposes to stabilize the error dynamics toward the demonstrated behavior. Similarly, \cite{LDM, nawaz2024learning} learn a policy that converges toward the training dataset by utilizing Lyapunov stability. Methods like \cite{perez2023stable, euclideanizing_flows} use diffeomorphic mappings to learn stable motions. While these methods show impressive performance, they rely on multiple demonstrations with extra training time. 

One-shot imitation learning has recently gained attention, exploring techniques such as self-attention and contrastive learning \cite{mandi2022onetowards}, invariant region matching \cite{zhang2024one}, and interaction warping \cite{biza2023one}. In contrast, we propose EMP as an alternative, emphasizing computational efficiency, converging behaviors, and real-time adaptability.

In our work, we are interested in both data and time efficiency. We focus on minimal demonstrations with minimal training time and real-time adaptation, following the stable Dynamical System (DS)-based motion policy \cite{TEXTBOOK, khansari2011learning, figueroa2018physically, se3lpvds}. With this direction, we would like to study: \textit{how much generalizability can be squeezed from simple and explainable models for stable imitation learning?} However, previous DS works focus on learning a fixed policy. While they can generate motion in areas that are not covered by data with Lyapunov stability constraints, the motion tends to become invalid with changes in the environment, requiring new demonstrations. Our prior work on Elastic-DS \cite{li2023task} proposes a solution by morphing the DS motion policy parameters based on the environment constraints. However, it is limited in Euclidean space with ground-truth targets given. In this work, we demonstrate the ability to extend the method to full pose with accelerated computational efficiency. We also show how our method can be combined with recent developments in pre-trained models to achieve more capabilities.

\section{BACKGROUND}
\label{sec:background}
The Elastic Motion Policy (EMP) is built on the SE(3) LPV-DS motion policy learning framework~\cite{se3lpvds}. We introduce the formulation and optimization of their parameters.
\vspace{-5pt}
\subsection{SE(3) Linear Parameter Varying Dynamical System}
To learn a DS motion policy in full end-effector space the SE(3) LPV-DS~\cite{se3lpvds} is formulated in the approximate $\mathbb{R}^3\times SO(3)$ space. The translational motion is learned with the classic LPV-DS framework~\cite{figueroa2018physically}, whereas the rotational motion is learned with a  Quaternion-DS formulation~\cite{se3lpvds}. 
\subsubsection{LPV-DS Formulation}
We begin by introducing the original LPV-DS framework, which is typically used to encode position trajectories. Let $x, \dot{x} \in \mathbb{R}^m$ represent the kinematic robot state and velocity vectors and $x^*$ be the attractor, the LPV-DS encodes a nonlinear DS as a mixture of continuous linear time-invariant (LTI) systems~\cite{figueroa2018physically,TEXTBOOK}:
\begin{equation} \label{eq:lpvds}
    \dot{x}=\sum_{k=1}^K \gamma_k(x)\bold{A}_k \left(x-x^*\right),
\end{equation}
where $K$ represents the total number of the LTI systems and $\gamma_k(x)$ is the state-dependent mixing function that quantifies the weight of each LTI system. $\gamma_k(x)$ is characterized by the GMM parameters $\Theta_{\gamma}=\{\pi_k, \mu_k, \bold{\Sigma}_k\}_{k=1}^K$ which are estimated by fitting a GMM to the reference trajectory~\cite{damm}. Subsequently, each LTI system $\bold{A}_k$ can then be learned by solving a semi-definite program (SDP) introduced in~\cite{figueroa2018physically} with constraints enforcing globally asymptotic stability (GAS)~\cite{Khalil:1173048} derived from a parametrized quadractic Lyapunov function (P-QLF) with the following form,
\begin{equation}
\label{eq:pqlf}
  V(x) = (x-x^*)^T\bold{P}(x-x^*)  
\end{equation}
with $\bold{P}=\bold{P}^T\succ0$ defining the elliptical shape of the Lyapunov function ~\cite{figueroa2018physically, TEXTBOOK}. Hence, the SDP minimizes the Mean Square Error (MSE) against the reference trajectories over the DS parameters $\Theta_{DS} = \{\bold{A}_k\}_{k=1}^K$:
\begin{equation}
\label{eq:lpvds_learning}
\begin{gathered}
    \min_{\Theta_{DS}} J\left(\Theta_{DS}\right)=\sum_{i=1}^{N}\left\|\dot{{x}}_{i}^{\mathrm{ref}}-{f}\left({x}_{i}^{\mathrm{ref}}\right)\right\|^2_2\\
    \text{s.t.} \,\,  \left\{\begin{array}{l}
    \left(\bold{A}_k\right)^T \bold{P}+\bold{P} \bold{A}_k=\bold{Q}_k\\ \bold{Q}_k=\left(\bold{Q}_k\right)^T \prec 0 \end{array}\right. \ \forall k=1,\dots,K
\end{gathered}
\end{equation}
\subsubsection{Quaternion Dynamical System}
The quaternion dynamical system, or Quaternion-DS, is formulated on the tangent plane of the quaternion space~\cite{se3lpvds}. For clarity, we denote elements of the manifold in bold and elements in tangent space in fraktur typeface; i.e., $\bold{q} \in \mathcal{M}$ and $\frak{q} \in T_\bold{q}\mathcal{M}$. Unlike LPV-DS, which outputs continuous velocity, Quaternion-DS generates the next desired orientation given the current quaternion $\bold{q}$ and attractor $\bold{q}_{att}$:
\begin{equation} \label{eq:quat_ds}
    (\Hat{\mathfrak{q}}_{att})^{des} = \sum_{k=1}^K \gamma_k(\bold{q}) \bold{A}_k\log_{\bold{q}_{att}} \bold{q},
\end{equation}
where the Riemannian logarithmic map $\log_{\bold{q}_{att}}\bold{q}$ computes the deviation between the current orientation $\bold{q}$ and the target $\bold{q}_{att}$~\cite{RIEM-COV,RIEM-STAT,RIEM-STAT-2}. The mixing function $\gamma_k(\bold{q})$ is defined similarly as its Euclidean counterpart except its parameters $\Theta_{\gamma} = \{\pi_k, \tilde{\mu}_k, \tilde{\bold{\Sigma}}_k\}_{k=1}^K$ are estimated by fitting a mixture model on the quaternion trajectory projected to the tangent space defined by $\bold{q}_{att}$~\cite{se3lpvds}. Each LTI system is then learned by solving a semi-definite program where the MSE is minimized on the tangent plane instead to preserve the Euclidean metric:
\begin{equation}\label{eq:quat_ds_learning}
\begin{aligned}
\min_{\Theta_{DS}} &J\left(\Theta_{DS}\right)=\sum_{i=1}^N \left\| (\Hat{\mathfrak{q}}_{att}^i)^{des}  -   (\mathfrak{q}_{att}^i)^{des} \right\|^2_2 \\
\text{s.t.}& \,\,  \bold{A}_k \prec 0, \quad \forall k = 1, \dots, K
\end{aligned}
\end{equation}
Similar to Eq.~\ref{eq:lpvds_learning}, the constraint enforces GAS using a quadratic Lyapunov function. Derivation found in~\cite{se3lpvds}. 


\subsubsection{Stable SE3-LPVDS Policy} Combining the LPV-DS for position control and Quaternion-DS for orientation control gives us a full pose motion policy with stability guarantees:
\begin{equation}
\begin{aligned}
        \Dot{x} = f_p(x;\ \Theta_p), \, \,\,\,\omega = f_o(\bold{q};\ \Theta_o)
\end{aligned}
\end{equation}
where each function is parameterized by $\Theta_{*} = \{\Theta_\gamma, \Theta_{DS}\}$ introduced from the previous subsections. Linear velocity $\Dot{x}$ follows the Eq.~\ref{eq:lpvds}. Computing angular velocity $\omega$ requires additional steps after obtaining the next desired quaternion in Eq.~\ref{eq:quat_ds}, involving parallel transport and Riemannian exponential map to recover the unit quaternion from the tangent plane (see Appendix~\ref{appendix:quat-ds} for details). 

\section{ELASTIC MOTION POLICY}
\label{sec:emp}
In this section, we present the EMP approach, which allows real-time adaptation of SE3-LPVDS to different environment configurations without new data. The key insight is to introduce geometric constraints to morph the learned SE3-LPVDS parameters based on their spatial changes. 
\vspace{-2.5pt}
\subsection{SE3 LPV-DS Policy Morphing}\label{sec:elastic-ds}
Elastic-DS \cite{li2023task} allows morphing an existing LPV-DS motion policy by posing geometric constraints on the GMM parameters $\Theta_\gamma$. The key idea is to leverage the geometric relationship between neighboring Gaussians in the GMM by specifying \textit{joints} $\beta_{i,k,k+1}$, a point in the same space as the policy, which is approximately a middle point between two neighboring Gaussians in the GMM,
\begin{equation}
\bold{\Sigma}_t=\left(\bold{\Sigma}_k^{-1}+\bold{\Sigma}_{k+1}^{-1}\right)^{-1} 
\end{equation}
\begin{equation}\label{eq:joint}
\beta_{i, k,k+1}=\bold{\Sigma}_t\left(\bold{\Sigma}_{k}^{-1} \mu_k+\bold{\Sigma}_{k+1}^{-1} \mu_{k+1}\right).
\end{equation}
where $\Sigma_{t}$ is the covariance of a distribution obtained by two neighboring Gaussians $\Sigma_k$ and $\Sigma_{k+1}$ obtained from DAMM \cite{damm}.
After that, by utilizing Laplacian Editing, 
\begin{equation}\label{eq:laplacian_editing}
\begin{aligned}
&\min _{\beta_i} J\left(\beta_i\right)=\left\|L \beta_i-\Delta\right\|_2^2 \\
&\text { s.t. } \,\left\{\begin{array}{l}
T_{0,1}\left(\beta_{i, 0}, \beta_{i, 1}\right)=O_{\text {start }} \\
T_{n-1, n}\left(\beta_{i, n-1}, \beta_{i, n}\right)=O_{\text {end }},
\end{array}\right.
\end{aligned}
\end{equation}
where $L$ is the graph Laplacian matrix, $\Delta$ is the Laplacian coordinates. Homogeneous transformation constraints related to relevant objects in the demonstrations $O_\text{start}$ and $O_\text{end}$ can be specified at the endpoints $T_{0,1}$ and $T_{n-1, n}$ while the other joints will adjust ``elastically" to preserve local geometric relationships. Hence the name Elastic-DS. Given the new GMM, the motion policy is re-estimated with Eq. \ref{eq:lpvds_learning}. 
For more details on the Euclidean Elastic-DS, refer to \cite{li2023task}.

Similarly, if we can extract the \textit{joints} $\beta_{i,k,k+1}$ in orientation space, we can apply the same operation as in Eq.~\ref{eq:laplacian_editing} to manipulate those \textit{joints} and adjust the original demonstration to different orientation behaviors. As introduced in~\cite{se3lpvds}, the mean and covariance of quaternion mixture model are defined on the tangent plane $T\mathbb{S}^{3}$, which is a 3-dimensional subspace embedded in $\mathbb{R}^4$. 
Since all the quaternion GMM means, $\Tilde{\mu}_k$, are expressed in terms of the attractor, we can form a basis centered at the attractor by defining the null space of the attractor vector in quaternion space, where the null space of a 4-dimensional vector is a 3-dimensional hyperplane embedded in $\mathbb{R}^4$: 
\begin{equation}
\text{Null}(\bold{q}_{att}) = \left\{ \bold{q}_{i} \in \mathbb{R}^4 \mid \bold{q}_{i} \cdot \bold{q}_{att} = 0, \ \forall i=1,2,3 \right\}.
\end{equation}
We then form the basis of the hyperplane as:
\begin{equation}
\bold{\Lambda}_{\bold{q}_{att}} = \begin{bmatrix} \bold{q}_{1}\ \bold{q}_{2}\ \bold{q}_{3}
\end{bmatrix}\in \mathbb{R}^{4\times 3}.
\end{equation}
to express the 4D mean vectors, $\Tilde{\mu}_k$, in 3D coordinates as:
\begin{equation}
\Hat{\mu}_k =  \bold{\Lambda}_{\bold{q}_{att}}^T \log_{\bold{q}_{att}}(\Tilde{\mu}_k) \in \mathbb{R}^3,\ \forall k=1,\dots, K.
\end{equation}
Since each covariance matrix, $\Tilde{\bold{\Sigma}}_k$, is defined wrt. $\Tilde{\mu}_k$ instead of the attractor, we form a set of the corresponding bases by finding the null space of each quaternion mean:
\begin{equation}
\left \{
\bold{\Lambda}_{\mu_k} = \begin{bmatrix} \bold{q}_{1}\ \bold{q}_{2}\ \bold{q}_{3}
\end{bmatrix} \mid \bold{q}_{i} \in \text{Null}(\Tilde{\mu}_{k}),\ \forall k=1,\dots, K
\right \},
\end{equation}
and we can then transform each covariance w.r.t. the corresponding basis to project them onto the 3D space:
\begin{equation}
\Hat{\boldsymbol{\Sigma}}_k = \bold{\Lambda}_{\mu_k}^T \Tilde{\boldsymbol{\Sigma}}_k \bold{\Lambda}_{\mu_k}\in \mathbb{R}^{3\times 3},\ \forall k=1,\dots, K
\end{equation}
We have now transformed all the quaternion terms into the 3D space and can perform the elastic transformation of orientation as in $\mathbb{R}^3$ following Eq.~\ref{eq:laplacian_editing}. We note that the geometric constraints, e.g., $\mathcal{O}_{\text{start}}$ and $\mathcal{O}_{\text{end}}$, should also be transformed and expressed in $\mathbb{R}^3$ wrt. the attractor. After the elastic transformation, we can obtain the newly transformed joints including new Gaussian mean and covariance. To recover the full mean and covariances, we project the reduced ones from $\mathbb{R}^3$ to the quaternion space as:
\begin{equation}
\Tilde{\mu}_k^* = \exp_{\bold{q}_{att}}( \bold{\Lambda}_{\bold{q}_{att}}\Hat{\mu}_k^*) \in \mathbb{H} \subset \mathbb{R}^4,
\end{equation}
and the covariance can be recovered as follows:
\begin{equation}
\Tilde{\boldsymbol{\Sigma}}_k^* = \bold{\Lambda}_{\mu_k}
\Hat{\boldsymbol{\Sigma}}_k^* \bold{\Lambda}_{\mu_k}^T\in \mathbb{R}^{4\times 4}.
\end{equation}

\begin{figure}[t!] 
    \centering
    \includegraphics[width=0.9\linewidth]{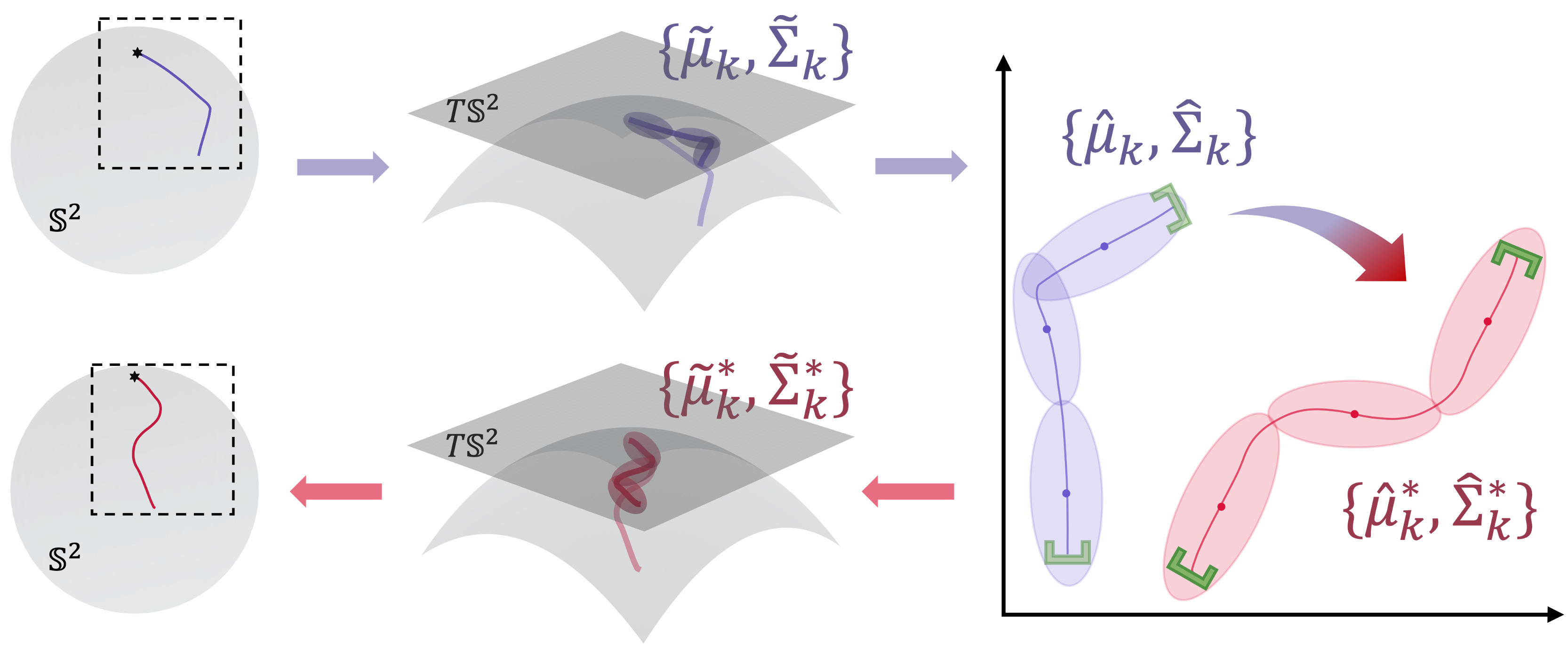}
    \caption{Illustration of elastic update of orientation trajectory, where the blue trajectory and its Gaussian parameters were first projected onto the tangent plane and then reduced to the lower dimensional Euclidean space. After update, the new trajectory in red and the updated Gaussians can be recovered by reversing the operation.}\label{fig:ori-update}
    \vspace{-15px}
\end{figure}

\begin{figure}[h!] 
    \centering
    \includegraphics[width=1\linewidth]{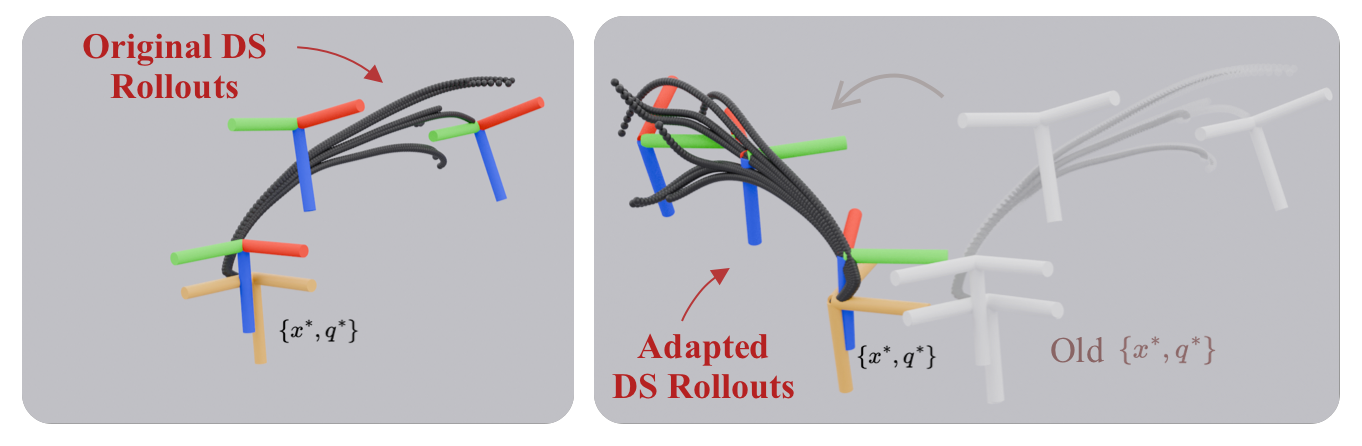}
    \caption{Illustration of elastic update of full pose trajectory.}\label{fig:full-update}
    \vspace{-15px}
\end{figure}

To illustrate the pipeline, we visualize the update of a toy trajectory in $\mathbb{S}^2$ in Fig.~\ref{fig:ori-update}. Once we have a new set of transformed Gaussians, i.e., $\{\Tilde{\mu}_k^*, \Tilde{\boldsymbol{\Sigma}}_k^*\}_{k=1}^K$, we can then feed the new Gaussians parameters into the learning of new Lyapunov functions and linear systems as introduced in next. Combining with Euclidean space Elastic-DS, Fig.~\ref{fig:full-update} shows an example of the full pose trajectory update.
\vspace{-10px}

\subsection{Real-Time Convex Lyapunov Function Optimization} \label{sec:convex_learning}
In the original LPV-DS \cite{figueroa2018physically} learning framework, 
the $\mathbf{P}$ matrix in Eq. \ref{eq:pqlf}
was estimated using a nonconvex optimization formulation from \cite{khansari2014learning}, which hinders the ability to update in real-time. Here, we propose to estimate $\mathbf{P}$ via a convex optimization by considering the demonstration data $\{x_{i}, \dot{x}_i\}_{i=1}^N$ and the unique attractor $x^{*}$. 
The goal is to estimate $\mathbf{P}$ by minimizing the number of training data points that violate the following Lyapunov stability condition,
\begin{align}\label{eq:vio}
    \Dot{V}(x) = \Dot{x} \cdot \nabla V  = 2\Dot{x}^T\bold{P}(x-x^*) < 0, \quad \forall \ x\neq x^*.
\end{align}

We then formulate the estimation of $\bold{P}$ as follows:
\begin{equation}\label{eq:convex_p}
\begin{array}{ll}
\underset{\bold{P}} \min & \sum_{i=1}^{N} \textbf{ ReLU}({\dot{x}}_{i}^T \mathbf{P} ({x}_{i} - x^*)) \\
\text { s.t. } & \mathbf{P} \succeq \epsilon I,
\end{array}
\end{equation}
where $\epsilon > 0$ is a small value to ensure numerical stability. Given that \textbf{ReLU} is a convex function, Eq. \ref{eq:convex_p} is a convex optimization problem \cite{boyd2004convex} that can be solved via QPs.

\begin{figure}[!h] 
\vspace{-5px}
    \centering
    \includegraphics[width=1\linewidth]{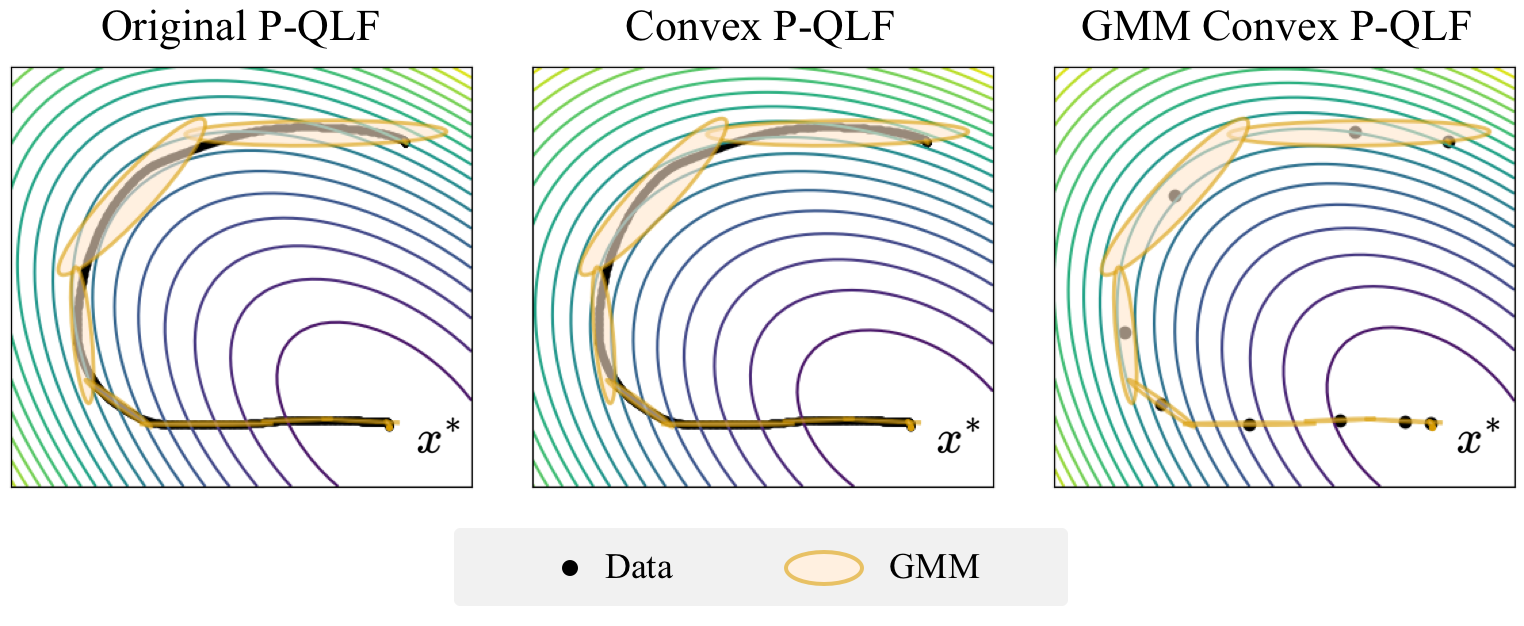}
    \caption{An example of fitting $\mathbf{P}$ (represented by the contour level set) using the original P-QLF (left), the newly proposed Convex formulation (middle), and the newly proposed GMM-informed convex formulation (right). Our methods speed up the optimization for real-time purposes while retaining performance.}\label{fig:convex_p_plot}
    \vspace{-15px}
\end{figure}

Besides the convex formulation, the proposed method can be further sped up in practice by taking advantage of the additional information from the statistical GMM model $\Theta_\gamma$. Since each Gaussian represents a region of neighboring data sharing similar characteristics, we can substitute the states of each data point in~Eq. \ref{eq:convex_p} with the average position and velocity of each Gaussian as shown in Fig. \ref{fig:convex_p_plot}. The simplified waypoint trajectory significantly reduces the computation overhead and enables EMP to update its policy in real time based on relevant object changes, as shown in Section \ref{sec:experiment}.


\begin{figure*}[htb!]
  \centering
  \includegraphics[width=0.85\textwidth]{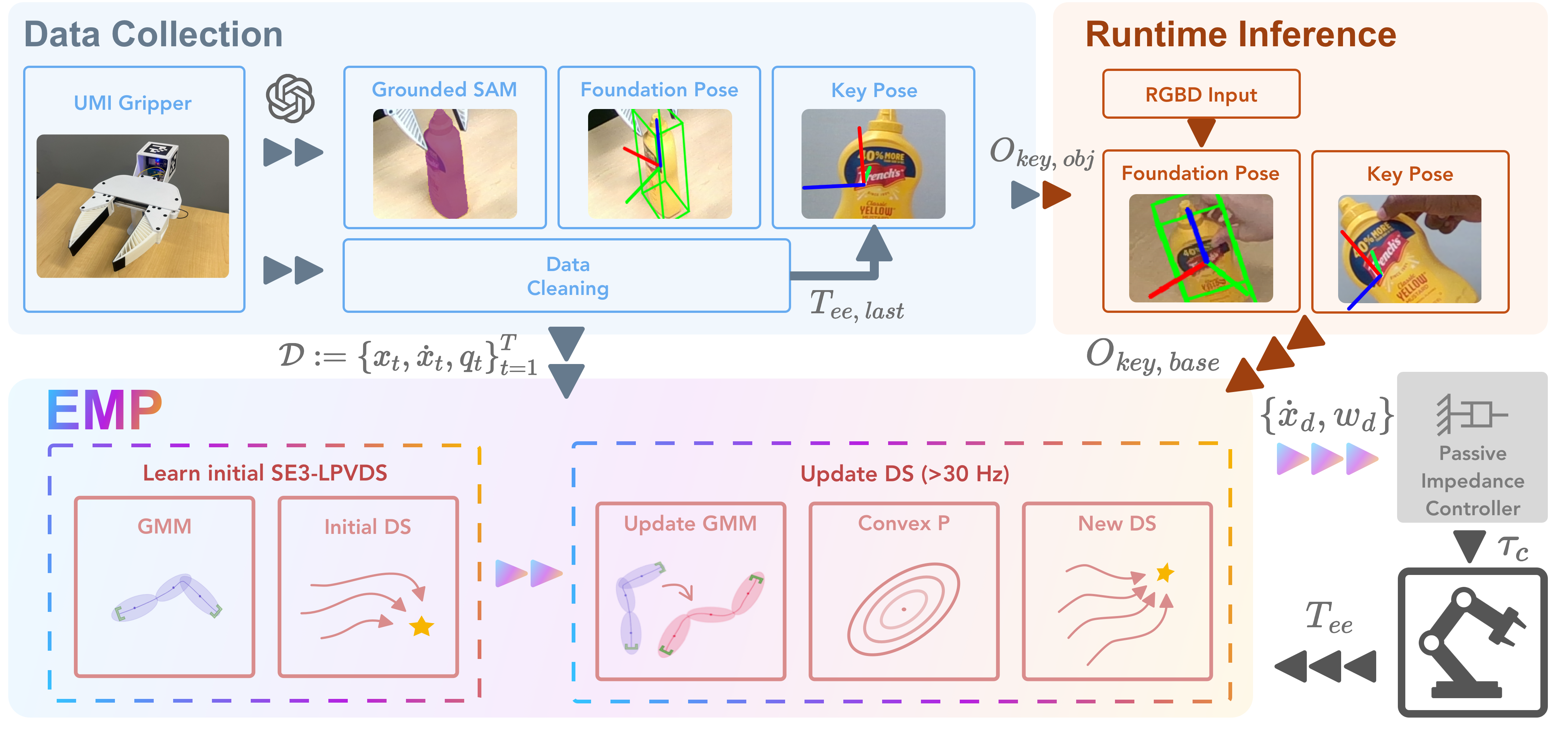}
  \vspace{-5pt}
  \caption{A block diagram showing the full EMP imitation learning pipeline.\label{fig:block-diagram}}
  \vspace{-15pt}
\end{figure*}
\section{EMP IMITATION LEARNING PIPELINE} 
\label{sec:pipeline}
\vspace{-2pt}
The full EMP system pipeline is illustrated in Fig. \ref{fig:block-diagram}.
\vspace{-5pt}
\subsection{Data Processing}
\subsubsection{Demonstration Data Collection}
We customized the UMI gripper~\cite{chi2024universal} to collect demonstration data, and use an external camera to record the demonstration in RGBD format, as shown in Fig. \ref{fig:block-diagram}. To alleviate the challenges of external camera placing and occlusion, multiple AprilTags \cite{olson2011apriltag} were strategically placed on different sides of the onboard UMI cube, allowing for a smooth and intuitive data collection process without requiring a robot. To determine the gripper states, we implemented a microcontroller-based contact sensor that records binary gripper states from the demonstration. Another AprilTag cube is placed as a pseudo-robot base during the demonstrations, used to convert all recorded trajectories $\{(x_i, \dot{x}_i, q_i)\}_{i= 0...n}$ into the robot base frame. During execution, the same reference Apriltag cube is attached to the robot to define the base frame.
\subsubsection{Keypose/Attractor Extraction} \label{sec:keypose}
\begin{figure}[h!]
    \centering
\includegraphics[width=0.8\linewidth]{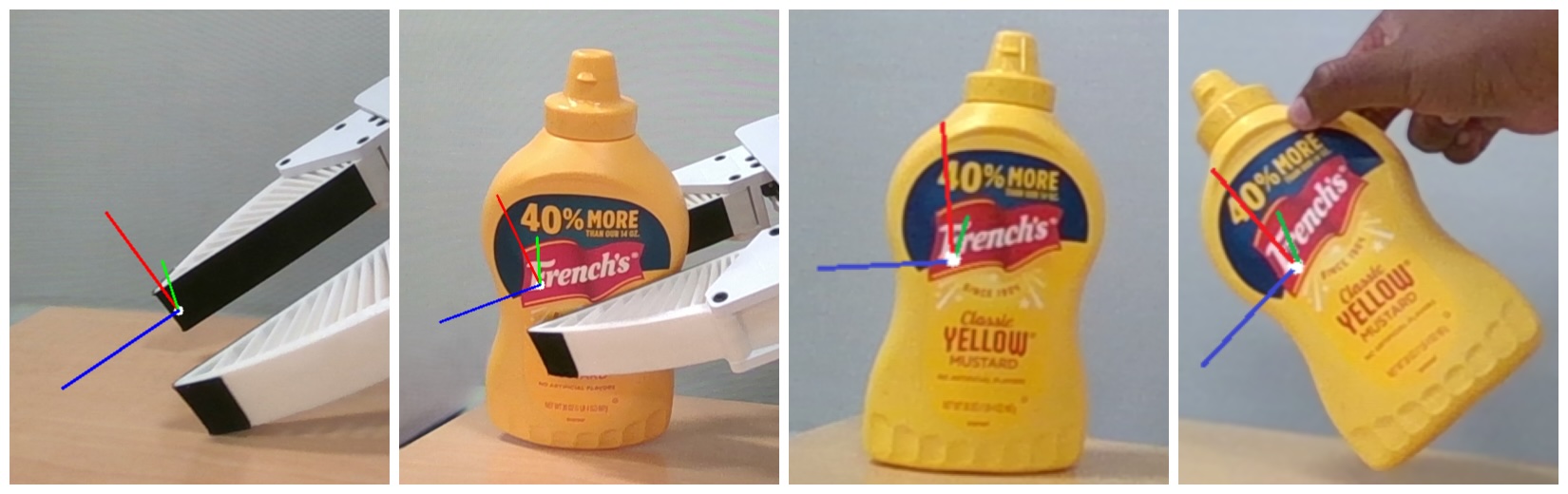}
    \caption{The first two images show the selection of the keypose using the UMI gripper end-effector fixed in the body frame of the object. This selected keypose is then tracked using FoundationPose~\cite{wen2024foundationpose}.\label{fig:keypose_annotation}}
    \vspace{-5pt}
\end{figure}

EMP learns a goal-oriented motion policy, requiring a final desired pose attractor. We define such a pose attractor using a \textit{keypose}, as shown in Fig.~\ref{fig:keypose_annotation} for a mustard bottle. When an object's pose changes, its \textit{keypose} will change and become the geometric constraint for the EMP adaptation in Eq. \ref{eq:laplacian_editing}. To retrieve the \textit{keypose} from demonstration, we use a language model to determine the semantic label from the video frames (first, middle, and final) for the relevant object related to the end pose $O_{ee, last}$. An example prompt for GPT-4o is shown in Appendix.~\ref{appendix:gpt}. The semantic phrase is then fed into Grounded SAM \cite{ren2024grounded} to generate the object mask. FoundationPose \cite{wen2024foundationpose} then takes the mask to determine the object pose $O_{obj} \in SE(3)$. The last end-effector pose $O_{ee, last}$ from the demonstration in the relevant object frame $O_{obj}$ will be recorded as the keypose $O_{key, obj}$. This representation is simple yet effective, and can represent keyposes for many tasks as shown in Section \ref{sec:experiment}. 

\subsubsection{Multi-Step Trajectory Decomposition} \label{sec:segmentation}
We divide a long-horizon demonstration into multiple sub-goal trajectories to learn individual goal-oriented motions. Then, we stitch multiple motion policies together to compose the multi-step task, $\dot{x}=\sum_{k}\,\sum_{j} \delta\left(\xi\right) f_{kj}\left(x\right),$ where $f_{kj}\left(x\right)$ represents each of the single stable motion policies, and $\delta(\xi)$ is a one-hot activation function that activates the next DS when the current one reaches its attractor. The outermost sum indexing by $k$ is the decomposition at the task subgoal level. Preliminary results have been shown in our prior Elastic-DS paper \cite{li2023task} with manually defined segments. In this work, we utilize pre-trained visual representation models, Universal Visual Decompose (UVD)\cite{zhang2024universal}, for the decomposition. This approach fits well with EMP as it can decompose long-horizon tasks into goal-conditioned subtasks. By applying UVD directly on the recorded demonstration video, it decomposes the trajectory, and each segment is then processed individually following the data processing pipeline. In this case, different segments could have different keypose for the same object.
\vspace{-5pt}
\subsection{Runtime Policy Inference}
After processing the demonstration, we train the nominal SE3-LPVDS policy. By tracking object poses $O_{obj}$, we update the EMP policy online using the keypose $O_{key, r}$ as a geometric constraint in Eq.~\ref{eq:laplacian_editing}. The output ${\dot{x}, w}$ is passed to a passive impedance controller \cite{kronander2015passive}, converting desired velocities to joint torques. For multi-step tasks, reaching a subtask attractor triggers the next subtask via the activation function $\delta(\xi)$, with gripper state changes occurring only at subtask completion.

\section{EVALUATION}
\label{sec:experiment}
In this section, we evaluate the proposed EMP approach from two different perspectives: 1) the performance of the convex P-QLF formulation benchmarked using the LASA 2D handwriting datasets, and 2) the efficacy of the \textit{elastic} adaptation tested on real-world tasks of varying complexity.
\subsection{Convex Lyapunov Function Evaluation}
The proposed convex optimization for P-QLF learning is evaluated on the LASA datasets \cite{khansari2011learning}, a collection of 30 human handwriting motions in 2D, each comprising 7 trajectories, with a total of 7000 observations. We compared two variants of our approach against the baseline formulation, using two metrics: 1) \textit{computation time} and 2) \textit{percentage of violation}, where the number of violations defined in Eq.~\ref{eq:vio} is divided by the total number of data points in a test trial. 

To align with the one-shot nature of our approach, we first evaluate it using a single trajectory. For each motion in the dataset, we randomly sample a single trajectory and compute the above metrics. We repeat the process 10 times per motion, and average the results across the entire dataset. As shown in Table.~\ref{tab:comparison}, both variants of our approach significantly outperform the baseline in computation time. The increase in violation for the GMM-informed convex P-QLF is expected, as using the average velocity within each Gaussian improves the speed at the expense of losing the detail of individual data points. Nevertheless, we achieve comparable or better results than the baseline. For completeness, we evaluate on all 7 trajectories for each motion in the dataset. All methods scale linearly with the increasing number of data points as shown in Table.~\ref{tab:comparison}. Nevertheless, the GMM-informed P-QLF still achieves a nearly 50 times improvement in computation speed compared to the baseline method.


\begin{table}[!tbp]
\scriptsize
\centering
\resizebox{\columnwidth}{!}{
\begin{tabular}{c|c|c|c}
\hline  \rule{0pt}{7pt}  
 LASA Dataset & Methods & Computation Time (s) & Violation Percentage\\ \hline \hline
 \rule{0pt}{6pt} 
\multirow{3}{1.2cm}{\centering \textbf{Single Trajectory}} & Baseline P-QLF & 0.332 & 14.0\% \\
 &Convex P-QLF &  0.038 & \textbf{11.1\%} \\
 &  GMM P-QLF & \textbf{0.007} & 15.1\% \\
 \hline   \hline
\rule{0pt}{6pt} 
\multirow{3}{1.2cm}{\centering \textbf{All Trajectory}} & Baseline P-QLF& 2.62 & 14.9\% \\
 &Convex P-QLF & 0.24 & \textbf{12.3\%} \\
 & GMM P-QLF & \textbf{0.09} & 15.4\%\\ \hline 
\end{tabular}
}
\caption{Comparison of different P-QLF learning formulations, with computation time measured in seconds. All experiments were conducted on an AMD Ryzen 7 5800X with 32GB of memory.\label{tab:comparison}}
\vspace{-10pt}
\end{table}

\begin{table}[t!]
\scriptsize
\centering
\resizebox{\columnwidth}{!}{
\begin{tabular}{cc|c|c|c}
\hline  \rule{0pt}{7pt}  
  & Methods & Book Placing & Cube Pouring & Pick-and-Place \\ \hline \hline
 \rule{0pt}{8pt} 
\multirow{2}{*}{\centering \textbf{ID}} & SE(3)-LPVDS & 10/10 & 10/10 & 8/10 \\
 &EMP (Ours) & 10/10 & 10/10 & 7/10\\
 \hline   \hline
\rule{0pt}{8pt} 
\multirow{2}{*}{\centering \textbf{OOD}} & SE(3)-LPVDS& 4/10 & 4/10 & 1/10 \\
 &EMP (Ours) & \textbf{8/10} & \textbf{9/10} & \textbf{7/10}\\
\hline
\end{tabular}
}
\caption{Success rates of all experiments evaluated in ID and OOD scenarios. Franka Research 3 robot arm was used.\label{tab:success_rate}}
\vspace{-20pt}
\end{table}
\subsection{Real Robot Tasks Evaluation}
We evaluate the \textit{elastic} adaptation of our proposed work on goal-oriented tasks: a) \textit{Book Placing}, b) \textit{Cube Pouring} and c) \textit{Pick-and-Place}, and additionally conduct a qualitative analysis of the success rate for both in-distribution (ID) and out-of-distribution (OOD) scenarios. Each task is demonstrated once, with starting and end points from the demonstration constituting the ID scenarios, while the OOD ones consist of random placement of the object and target. We choose to use the object-centric version of SE(3)-LPVDS~\cite{se3lpvds} as the baseline, where the learned DS policy is globally transformed based on the change in object keypose. The rest of the experiment setup follows the pipeline in Section.~\ref{sec:pipeline},
\begin{figure}[!tbp]
    \centering
    \begin{minipage}{\linewidth}
    \centering
  \includegraphics[width=\columnwidth]{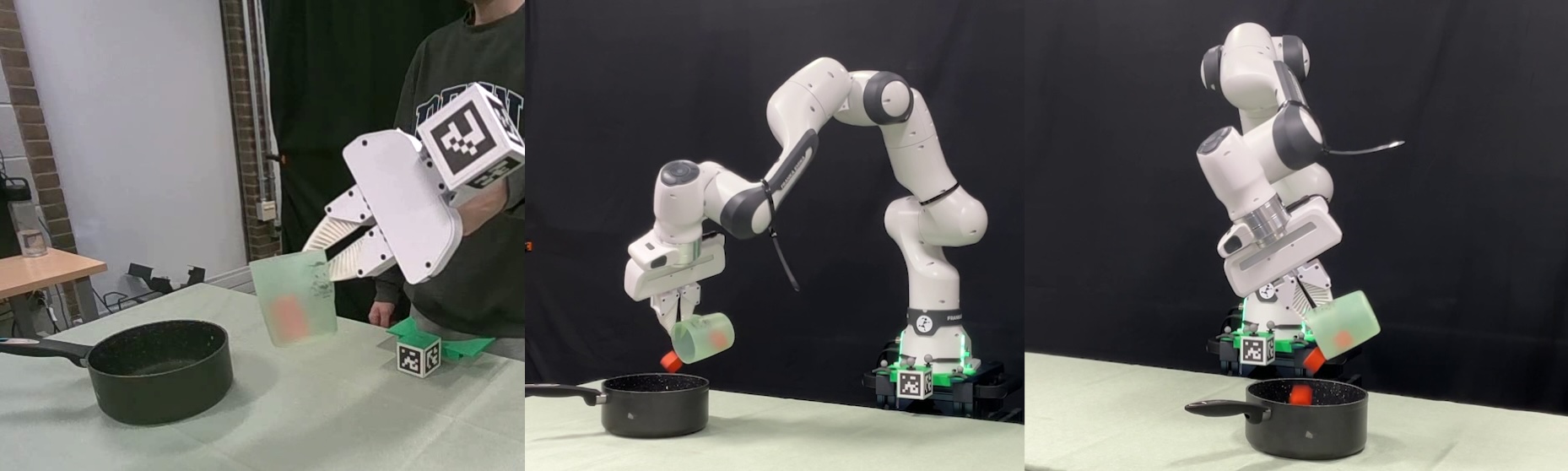}
  \caption{Given a single demonstration of cube pouring shown in the left, EMP can reproduce the same motion and generalize to different pot poses shown in the middle and right.\label{fig:pouring_task}}
    \end{minipage}
    \begin{minipage}{\linewidth}
    \centering
      \includegraphics[width=\columnwidth]{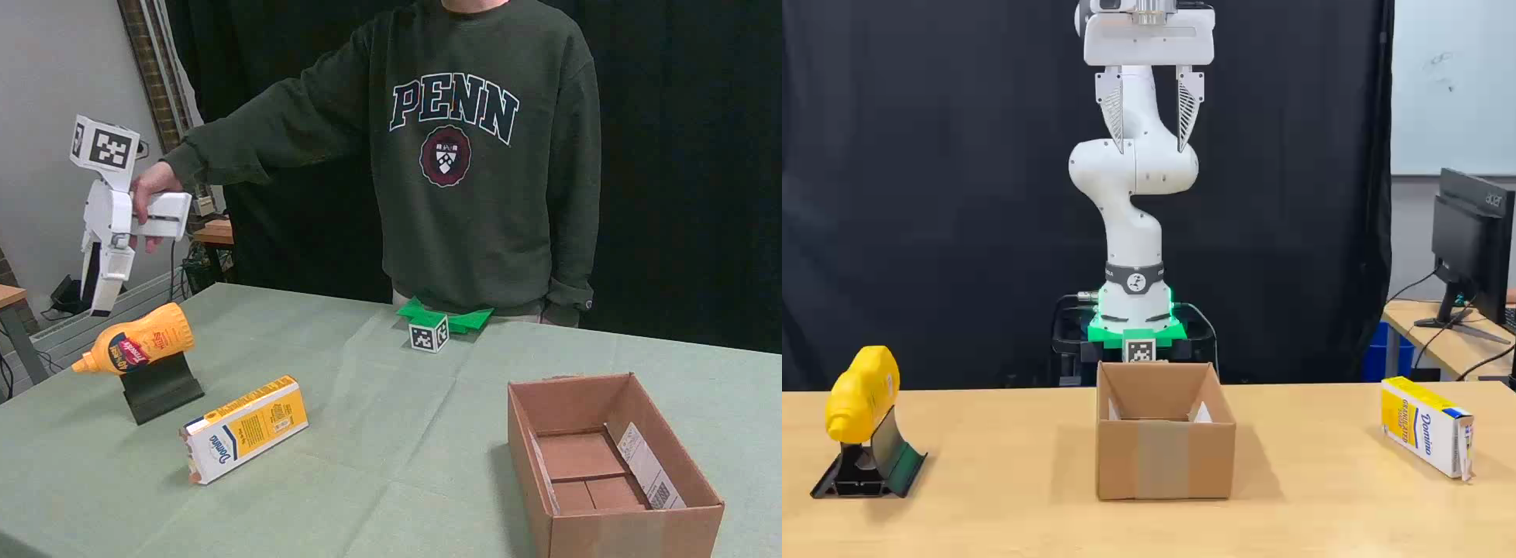}
  \caption{The multi-step task requires sequential pick and place motions. The demonstration has both objects on one side and the box on the other. During inference time, the box is placed in between the two objects.\label{fig:pnp_task}}
    \end{minipage}
    \begin{minipage}{\linewidth}
    \centering
      \includegraphics[width=\columnwidth]{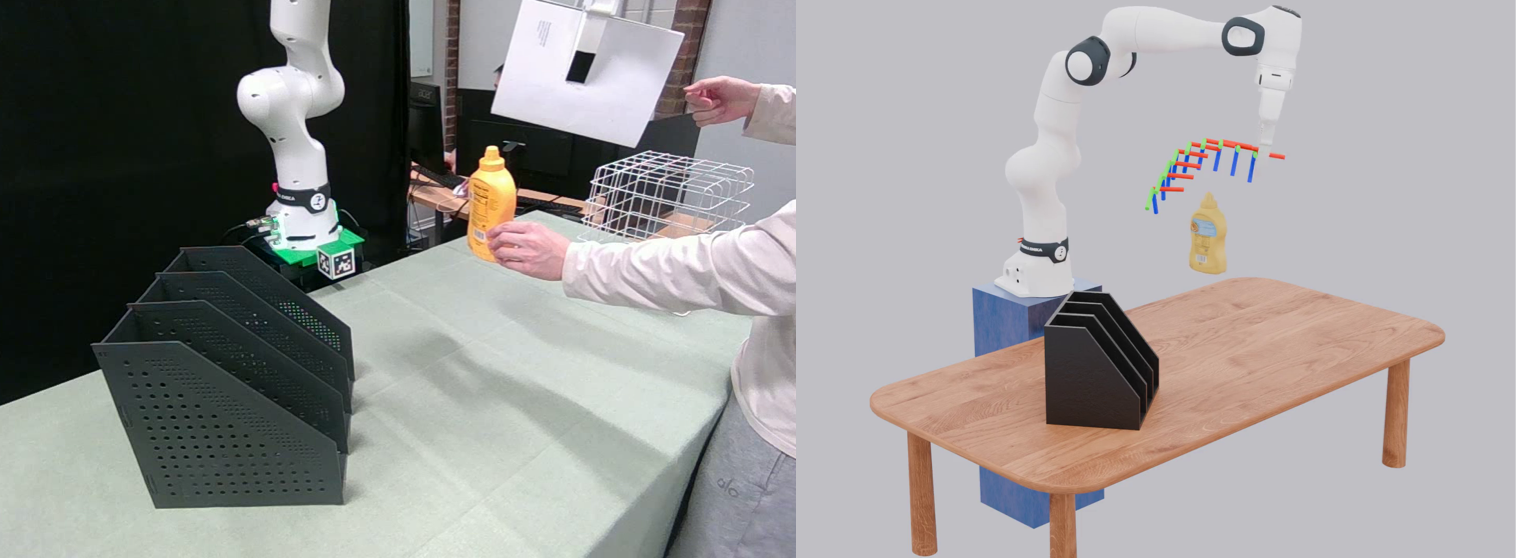}
  \caption{In the same \textit{Book Placement} task, a yellow bottle obstacle is blocking the original path. Modulation allows EMP to circumvent the obstacle, generating a new path above the bottle.\label{fig:obs_task}}
  \vspace{-20pt}
\end{minipage}
\end{figure}

\subsubsection{Book Placing}
For this task, we provide a single demonstration as shown in Fig.~\ref{fig:overview}, where the human demonstrator holds the book approaching the bookrack from the front opening and places it into the rightmost slot. Prior to execution, both the baseline method SE(3)-LPVDS and our approach learn the policy from the single nominal demonstration we provide. Both approaches then adapt their learned policy based on the scene changes defined by the keypose. We measure the success rate of the task completion over 10 runs each in both ID and OOD scenarios. In Table.~\ref{tab:success_rate}, we show that both methods succeed in the task when starting and ending are the same as the demonstration, i.e., ID scenarios. However, when we change the bookrack configurations more randomly and aggressively, the success rate of the baseline drops by half, while our approach manages to complete in more challenging OOD scenarios. We note that the specific constraints of this task come from both the placement pose and the approaching pose. For example, if the approaching pose is off, the book will collide with the bookrack and fail the task. By simply transforming (e.g. rotating and translating) a learned policy as in the baseline method, there is no guarantee that the task constraints are still valid in the transformed policy. On the other hand, EMP ensures that task constraints comply with the updated policy as enforced by Eq.~\ref{eq:laplacian_editing}, leading to higher success rate. An example of adaptation is shown in Fig.~\ref{fig:overview}.
\subsubsection{Cube Pouring}
In this task, the robot needs to pour a cup of cubes into a pot, as illustrated in Fig.~\ref{fig:pouring_task}. Between each trial, the cup always needs to recover back to the initial pose $O_{start}$ with its opening facing up, which is then used as the task constraint in Eq.~\ref{eq:laplacian_editing}. We first compare our approach against the baseline in the ID setting. Table.~\ref{tab:success_rate} shows that both methods achieve a perfect success rate due to the stability guarantees. However, in OOD scenarios, we observe that the baseline fails more than half of the trials. The constraints of this task require the robot to make an arch-shaped trajectory with wrist rotation for pouring, as in the original demonstration. However, the single demonstration only shows the motion approaching from one side of the pot. In one of the OOD cases, where the robot starts from the other side of the plot, it will enter an unknown area where the motion is less defined. Although the stability guarantee will ensure that the final pose goal is reached, the motion in between is unknown, which leads to behavior like generating a big circular motion or going toward the table. On the other hand, EMP remains robust in adapting to unforeseen configurations while ensuring that the motion complies with the demonstration due to the property of the Laplacian editing. The cause of the one failed case comes from the more challenging angle at which to pour the cubes. EMP adapts in task space and hence remains susceptible to situations where the adapted path--while satisfying the task constraints--may become infeasible for the robot, such as exceeding its reachability or joint limits. 
\subsubsection{Multi-Step Pick-and-Place Task}
This multi-step task requires the robot to continuously pick and place multiple objects into a packing box. A snapshot of the demonstration is shown on the left in Fig.~\ref{fig:pnp_task}. As introduced in Section.~\ref{sec:segmentation}, the entire task is decomposed into a sequence of single-segment goal-oriented tasks, with the UVD hyper-parameter $\gamma = 0.003$. The baseline method exhibits a drastic drop in performance between ID and OOD. Our proposed method also experiences more uncertainty in both settings, whereas EMP performs similarly. The lower performance of both the baseline and EMP on this task is due to inaccuracies in the segmentation and grasping failures.
For example, if the segmentation algorithm cuts the duration of placing motion too early--before the object is positioned above the box--the resulting keypose may then be misplaced outside the box. This misalignment may cause the robot to converge to an incorrect attractor and ultimately fail the task.
\subsection{Case Study: Book Placing with Obstacle Avoidance}
We demonstrate the additional feature of EMP to perform obstacle avoidance in real time, ensuring safe execution in a dynamic environment. We reuse the same experiment setup as in the \textit{Book Placing} task, and additionally place an mustard bottle as the obstacle to obstruct the path of booking placing motion. For simplicity, we model the obstacle as a sphere object with a $0.18m$ diameter. We can morph the learned velocity field by adopting the modulation approach~\cite{khansari2012dynamical}: $\dot{x}_{new} =\mathbf{M}(x) f(x)$, where $\dot{x}_{new}$ is the new velocity that incorporates the obstacle avoidance behavior and the modulation matrix $\mathbf{M}$ is constructed through eigenvalue decomposition with the normal and tangent directions of the obstacle boundaries. As shown in Fig.~\ref{fig:obs_task}, we demonstrate that EMP's compatibility with obstacle avoidance allows for more flexible behavior in real world tasks.

\section{CONCLUSIONS \& DISCUSSIONS}
In this work, we present EMP, a full-pose one-shot imitation learning approach with stability guarantees that can adapt to novel environment configurations in real-time. Through various experiments, we show how EMP can be naturally combined with other components, such as obstacle avoidance and multi-segment, to empower DS-based imitation learning. There are a few limitations and opportunities worth further exploration: (1) Our approach relies on accurate object tracking. In the future, we would like to explore using vision input directly to EMP (2). While the experiments specifically show capability in learning from a single demonstration, extending to multiple demonstrations is possible. By using DAMM \cite{damm}, we can fit the multi-demonstration with a single target pose as a directed graph. The general Laplacian Editing can be used for graphs, so it would be suitable for morphing such a structure as well. To accommodate multi-demonstration with multiple goals, where different locally specific motions are needed for different parts of the objects, the learned GMM can adapt based on the local geometric constraints. However, information about the object shape is then needed. (3) In our evaluations, we did not observe significant distortions using a single tangent space to approximate the global quaternion manifold structure, as highlighted by \cite{single-tangent-fallacy}. This stability may partly be attributed to the conservative nature of the quaternion DS constrained by a QLF. Future work will explore Lie group formulations that naturally respect the manifold structure. Furthermore, while the P-QLF efficiently addresses a broad class of reaching motions, it inherently restricts more complex, highly non-linear motions. For these complex cases, future directions include exploring more expressive formulations such as Neural-ODE approaches \cite{nawaz2024learning} or leveraging contraction theory. The above limitations provide further exploration opportunities to enrich the EMP framework. We believe that EMP provides a way towards learning adaptive, robust, and time and data-efficient motion.

\appendix
\subsection{Quaternion-DS Formulas}
\label{appendix:quat-ds}
Provided that a time difference $dt$ is known, we first parallel transport the estimated $(\Hat{\mathfrak{q}}_{att})^{des}$ obtained in Eq.~\ref{eq:quat_ds} from the attractor $\bold{q}_{att}$ back to the current state $\bold{q}$, 
\begin{equation}
        (\Hat{\mathfrak{q}}_{body})^{des}  = \Gamma_{\bold{q}_{att} \rightarrow \bold{q}}(\Hat{\mathfrak{q}}_{att})^{des},
\end{equation}
where the new vector is the estimated desired displacement expressed in the body frame. We then perform the Riemannian exponential map to project this vector from the tangent space back to the quaternion space, 
\begin{equation}
        (\Hat{\bold{q}})^{des} = \exp_{\bold{q}}{(\Hat{\mathfrak{q}}_{body})^{des}},
\end{equation}
where $(\Hat{\bold{q}})^{des}$ is the estimated desired orientation in quaternion space. We then compute the desired angular velocity,
\begin{equation}
        \omega = (\Bar{\bold{q}} \circ (\Hat{\bold{q}})^{des}) / dt.
\end{equation}
\vspace{-5pt}
\subsection{GPT Prompt for Semantic Object Extraction}\label{appendix:gpt}
We designed a prompt for GPT-4o to determine the relevant object in the demonstration video. The prompt comprises a series of sample images from the demonstration with text input. The output of GPT-4o serves as the prompt for Grounded-SAM. 
\begin{figure}[h!]
  \centering 
  \includegraphics[width=0.85\linewidth]{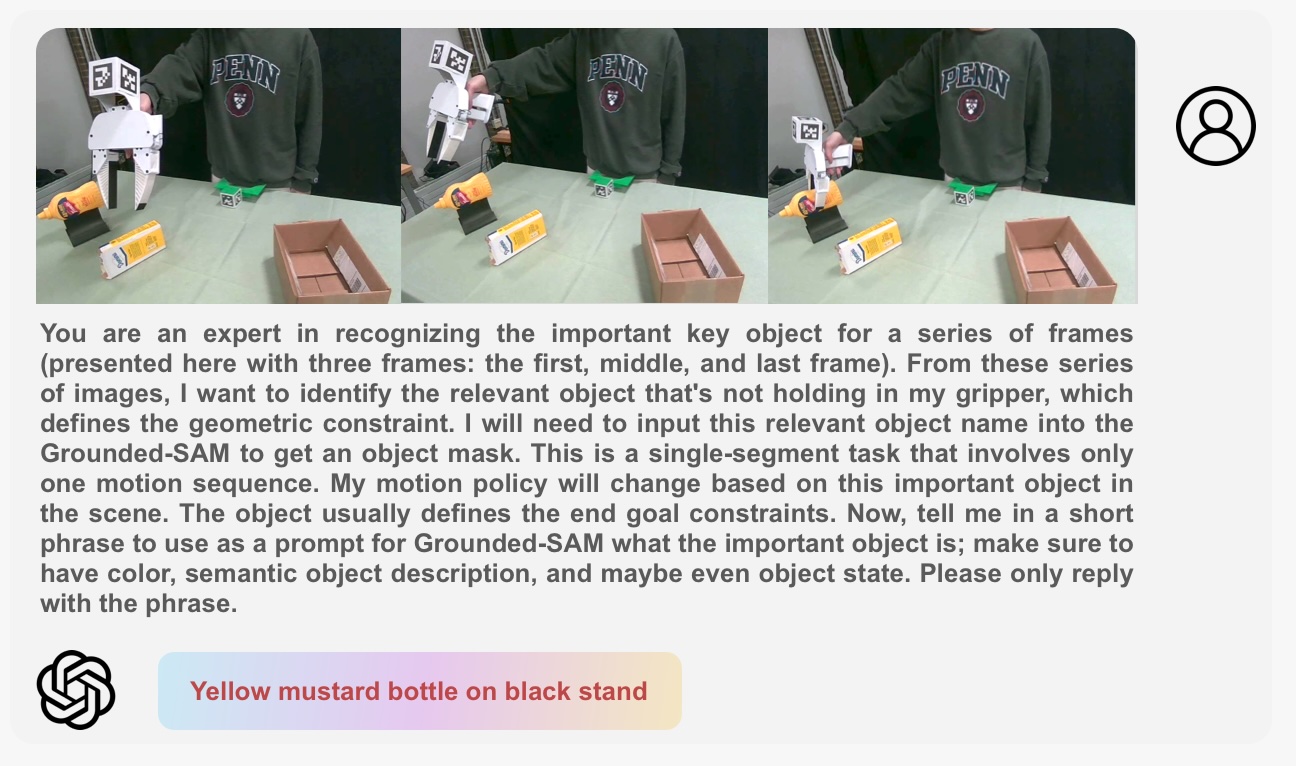}
  \caption{Prompt and response from GPT-4o\label{fig:gpt-prompt}}
  \vspace{-15pt}
\end{figure}

\bibliographystyle{IEEEtran}
\bibliography{root}

\end{document}